\documentclass[letterpaper]{article} 
\usepackage[preprint]{aaai2027}  
\usepackage[hyphens]{url}  
\usepackage{graphicx} 
\usepackage{natbib}  
\usepackage{caption} 
\usepackage{subcaption}
\usepackage{algorithm}
\usepackage{algorithmic}
\usepackage{pifont}

\usepackage{newfloat}
\usepackage{listings}
\DeclareCaptionStyle{ruled}{labelfont=normalfont,labelsep=colon,strut=off} 
\floatstyle{ruled}
\newfloat{listing}{tb}{lst}{}
\floatname{listing}{Listing}

\usepackage{multirow}
\usepackage{amsmath}
\usepackage{amssymb}

\usepackage{booktabs}

\title{PATH-Bench: Path-Dependent Evaluation of Lifelong Agents}
\author {
    Xidong Yang\textsuperscript{\rm 1}\equalcontrib\corresponding,
    Xingyi Zhang\textsuperscript{\rm 1}\equalcontrib,
    Wenhao Li\textsuperscript{\rm 2},
    Wenyan Liu\textsuperscript{\rm 3},
    Junjie Sheng\textsuperscript{\rm 4},
    Yun Hua\textsuperscript{\rm 5},
    Wei Yin\textsuperscript{\rm 6},
    Tao Fang\textsuperscript{\rm 3},
    Chuyun Shen\textsuperscript{\rm 7},
    Xiangfeng Wang\textsuperscript{\rm 1,\rm 8}\corresponding
}
\affiliations {
    \textsuperscript{\rm 1}East China Normal University\\
    \textsuperscript{\rm 2}Tongji University\\
    \textsuperscript{\rm 3}Ant Group\\
    \textsuperscript{\rm 4}Independent Researcher\\
    \textsuperscript{\rm 5}Shanghai Jiao Tong University\\
    \textsuperscript{\rm 6}Bank of Communications\\
    \textsuperscript{\rm 7}Shanghai University of International Business and Economics\\
    \textsuperscript{\rm 8}Shenzhen Loop Area Institute\\
    xdyang@stu.ecnu.edu.cn, xfwang@cs.ecnu.edu.cn
}

\begin{document}

\maketitle

\begin{abstract}
Lifelong LLM agents increasingly adapt through external learning states that store past interactions as retrievable memories or reusable skills, yet existing benchmarks rarely account for how the path of accumulated experience shapes what agents transfer and retain.
In this work, we establish \textbf{PATH-Bench}, a benchmark for path-dependent evaluation of lifelong agents.
PATH-Bench estimates directed task relationships via multi-model in-context learning and constructs probe-centered sequences with controlled helpful and interfering histories, scoring all tasks for average performance while tracking recurring probes to measure forward transfer, backward transfer, and forgetting.
We evaluate eight representative agents on single-turn code generation and multi-turn tool-use tasks under positive- and negative-dominant histories.
Benchmark results show that experience utility depends jointly on how experience is represented and on the task's interaction structure, that strong transfer does not ensure retention, and that later experience can reshape gains acquired earlier in the learning path.
Based on these findings, we propose Selective Experience Use (SEU), an agent harness that regulates how path-accumulated experience influences each new task, admitting helpful items while filtering out potential interference. SEU consistently reduces forgetting while improving forward transfer in the majority of settings.
PATH-Bench provides both a controlled evaluation framework and actionable guidance for designing more selective and robust lifelong agents.
The benchmark data and evaluation code are available at \url{https://github.com/Xidong-Yang/Pathbench}.
\end{abstract}


\section{Introduction}

Across the AI industry, a growing vision is to develop persistent agents that learn from experience, build memories, and improve over time across repeated interactions~\cite{openai2026frontier}.
This motivates lifelong LLM agents that accumulate and preserve useful experience over time~\cite{zheng2026lifelong,asawa2026continual,xu2026evoarena}. 
One approach is to internalize such experience through continual parameter updates, but doing so at LLM scale is costly and risks catastrophic forgetting~\cite{li2024revisiting,zheng2025towards,shi2025continual}. 
To avoid these limitations, many lifelong agents keep the backbone frozen and instead realize adaptation through the agent harness, which maintains an external learning state by storing past interactions as retrievable memories or reusable skills~\cite{hu2025memory,yang2026autoskill,yang2026agentic}.
However, their effectiveness depends on what is retained, how it is represented, and the path through which experience accumulates.

\begin{figure}[t]
    \centering
    \includegraphics[width=0.92\linewidth]{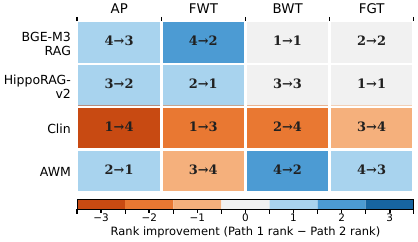}
    \caption{Agent-rank changes across two task paths on LifelongAgentBench~\cite{zheng2025lifelongagentbench}. Each cell shows the transition from Path 1 to Path 2 for average performance (AP), forward transfer (FWT), backward transfer (BWT), and forgetting (FGT); color encodes its direction and magnitude.}
    \label{fig:motivation_order}
\end{figure}

Despite its central role in continual learning~\cite{bell2022effect,benavides2022theory}, path dependence remains largely overlooked in evaluations of lifelong LLM agents.
Specifically, existing benchmarks do not control the task histories over which the agent harness updates its external learning state before evaluating later tasks~\cite{wei2025evo,asawa2026continual,xu2026evoarena}.
Positive experience may enable positive transfer, whereas misleading memories, outdated strategies, or conflicting skills may induce persistent interference.
To probe this sensitivity, we conduct a preliminary study on LifelongAgentBench~\cite{zheng2025lifelongagentbench}, the first unified benchmark for systematically assessing LLM agents as lifelong learners, and find that different task paths over an identical task set can change the relative ranking of agents (Figure~\ref{fig:motivation_order}; see Appendix~\ref{app:motivation_order} for details).
Consequently, evaluations over fixed or randomized paths cannot reveal how the accumulation of prior experience shapes later performance~\cite{baker2023domain}.

Characterizing this dependence requires tracing performance along controlled histories with metrics such as forward transfer, backward transfer, and forgetting~\cite{zuffer2025advancements,wolczyk2021continual}.
However, applying these analyses at scale to LLM agents is computationally challenging.
Evaluating a single agent task may already involve multi-step reasoning, tool use, environment interaction, and feedback~\cite{he2025llm,hu2024self}, while traditional lifelong learning metrics require repeatedly reevaluating the full task set throughout the learning trajectory~\cite{pan2025survey}.
The key challenge is to \textit{efficiently evaluate how task path dependence shapes the lifelong performance of LLM agents}, which is essential for assessing their ability to learn and adapt reliably from accumulated experience.

To address this challenge, we introduce \textbf{PATH-Bench}, a benchmark for path-dependent evaluation of lifelong agents.
The core idea is to replace exhaustive enumeration of task histories with sampling guided by task relationships, and to replace repeated evaluation of the full task set with longitudinal tracking of a single probe.
This design enables PATH-Bench to efficiently evaluate how task path dependence shapes the lifelong performance of LLM agents.

Across eight representative lifelong agents spanning retrieval-augmented generation (RAG), agentic memory, and skill-based approaches, we find that accumulated experience does not consistently improve performance, and its utility depends on whether the representation of experience matches the task's interaction structure.
We further observe a dissociation between transfer and retention, as strong immediate transfer can coexist with substantial forgetting.
Controlled path-transition experiments further show that later experience can reshape gains acquired earlier in the learning path.
Guided by these findings, we propose \textit{Selective Experience Use} (SEU), an agent harness that regulates how path-accumulated experience influences each new task, admitting helpful items in concrete or abstracted form while filtering out potential interference.
SEU consistently reduces forgetting across agents, datasets, and transfer histories, while improving forward transfer in the majority of evaluated settings, showing how PATH-Bench diagnostics can guide the design of lifelong agent harnesses that regulate the use of accumulated experience.

Overall, the main contributions are summarized as follows:
1) PATH-Bench is introduced, to the best of our knowledge the first benchmark for tracing transfer and retention in LLM agents under controlled histories: it estimates task relations with multi-model in-context learning (ICL), samples configurable relation-constrained sequences, and quantifies the resulting performance changes;
2) A controlled empirical study of eight representative lifelong agents is conducted, disentangling how the external learning state, its representation, and the learning path jointly shape transfer and retention;
3) Guided by these findings, we propose SEU, an agent harness that regulates how path-accumulated experience influences each new task, reducing forgetting and improving forward transfer in the majority of evaluated settings.

\begin{figure*}[t]
    \centering
    \includegraphics[width=\linewidth]{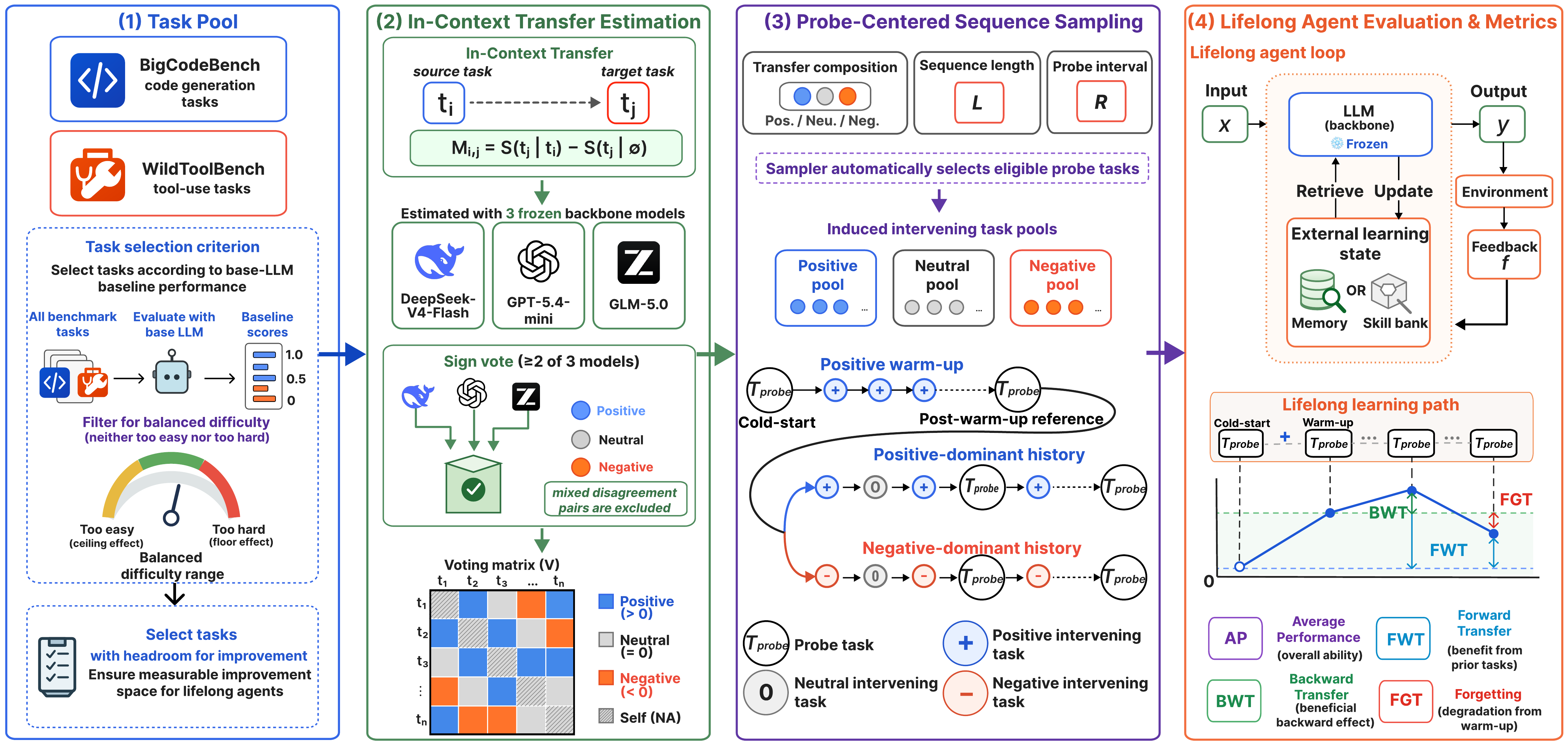}
    \caption{Overview of PATH-Bench.
    The framework filters both task pools for measurable headroom, estimates directed task relationships with a multi-model in-context transfer gain matrix, samples probe-centered sequences under positive- and negative-dominant histories, and evaluates agents that adapt through an external learning state with average performance, forward transfer, backward transfer, and forgetting.}
    \label{fig:overview}
\end{figure*}

\section{Related Work}

\noindent {\bf{Lifelong Learning}}.
Lifelong learning studies how systems acquire knowledge from task streams while balancing plasticity and stability, commonly analyzed through knowledge transfer and forgetting~\cite{parisi2019continual,wang2024comprehensive,lopez2017gradient}.
Classical approaches rely on regularization, experience replay, or parameter isolation~\cite{kirkpatrick2017overcoming,ye2023continual,ye2022lifelong}, but frequent online updates to large language models are costly and may destabilize their general capabilities~\cite{luo2025empirical}.
LLM agents adapt by evolving external states—memories, workflows, and skills—around a frozen backbone~\cite{park2023generative,zhang2026memrl,yang2026autoskill,ma2026skillclaw}.

\smallskip
\noindent {\bf{Evolution of LLM Agents}}.
Memory and skills are two primary substrates for experience-driven agent evolution~\cite{zhong2024memorybank,sumers2024cognitive}.
Memory methods differ in representation and update strategy: AWM~\cite{wang2024agent} extracts reusable workflows, Clin~\cite{majumder2023clin} distills causal rules, SimpleMem~\cite{liu2026simplemem} compresses interactions into structured guidance, and MemRL~\cite{zhang2026memrl} updates memory utility from environmental rewards; broader frameworks study memory formation, revision, and retrieval over time~\cite{zhang2025learn,li2025retrieval,du2025memr,cai2025flex}.
Skill-based agents abstract trajectories into reusable skills that can be discovered, routed, and refined without modifying the backbone~\cite{yang2026autoskill,ma2026skillclaw,alzubi2026evoskill,zhang2026memskill}.

\smallskip
\noindent {\bf{Lifelong Learning Benchmarks}}.
Existing LLM-agent lifelong learning benchmarks span three related directions.
Long-horizon memory benchmarks evaluate whether models retain and integrate information distributed across dialogues or agent-environment trajectories~\cite{maharana2024evaluating,castillo2024beyond,wan2025storybench,zhao2026amabench}.
Sequential adaptation benchmarks test whether agents improve from feedback streams, using performance trajectories, transfer, retention, or gains beyond underlying model capability~\cite{wu2024streambench,zheng2025lifelongagentbench,ai2025memorybench,asawa2026continual}, with EdgeBench further scaling this setting to ultra-long real-world tasks that accumulate environment interaction~\cite{zhu2026edgebench}.
Dynamic and self-evolving benchmarks further study environment changes, memory updates, and capability erosion across multiple forms of agent state~\cite{xu2026evoarena,yu2026capabilityerosion}.
Despite this breadth, existing benchmarks generally can not measure how task path dependence shapes both transfer and retention, while PATH-Bench is designed to close the gap.

\section{The PATH-Bench Benchmark}

The overview framework of PATH-Bench is provided in Figure~\ref{fig:overview}.
PATH-Bench uses an ICL-based transfer gain matrix to estimate directed task relationships and construct probe-centered sequences with configurable transfer patterns. By repeatedly evaluating probes under controlled histories, it measures average sequence performance and characterizes learning dynamics through forward transfer, backward transfer, and forgetting.

\subsection{Problem Formulation}

Consider an LLM agent that solves a stream of tasks while accumulating experience from the tasks it has processed.
Its behavior on any task therefore depends not only on that task but on the ordered history of tasks encountered before it.

Let $q$ be a fixed \emph{probe} task, $H=(h_1,\dots,h_k)$ an ordered history of intervening tasks that the agent processes before $q$, and $\mathcal{A}_H$ denotes the agent after processing $H$.
The expected performance of $\mathcal{A}_H$ on $q$ is defined as:
\begin{equation}
P_q(H)={\mathbb{E}}\bigl[\rho(y,q)\mid \mathcal{A}_H\bigr],
\label{eq:probe_score}
\end{equation}where $y$ is the agent's response to $q$ and $\rho(y,q)$ scores its success.
Performance is \emph{path-dependent} when $P_q(H)$ varies with the content and order of $H$.
In a lifelong sequence, the $i$-th occurrence of the probe sees a history prefix $H_i$, yielding the trajectory $p_i^{\mathrm{probe}}=P_q(H_i)$.
PATH-Bench evaluates an agent by contrasting this probe trajectory under controlled histories.


\subsection{Benchmark Construction Pipeline}
\label{sec:data_prep}

\noindent {\bf{Task Selection}}.
To evaluate the lifelong learning capabilities of LLM agents across distinct capability domains, we select two representative public benchmarks with substantially different task characteristics.
For code generation, we employ \textbf{BigCodeBench}~\cite{zhuobigcodebench}, a single-turn benchmark that evaluates practical program synthesis under complex instructions and diverse function calls, scoring each task by whether it is completed.
For tool use, we select \textbf{WildToolBench}~\cite{yubenchmarking}, a multi-turn benchmark grounded in real-world user behavior and targeting planning and execution on complex, cross-tool tasks, scoring each task by the proportion of successfully completed subtasks.
To ensure that lifelong agents have a meaningful opportunity to improve, we follow the headroom principle in continual learning benchmark design~\cite{asawa2026continual}: we estimate the baseline performance of strong LLMs on candidate tasks and filter out those already saturated or offering little observable learning signal.
From the resulting headroom-aware pool, we sample $D=120$ tasks from each benchmark to form the evaluation pool.

\noindent {\bf{Task Relationship}}.
Existing agent evaluations often rely on randomized task sequences that ignore inter-task relationships, making it difficult to study how earlier experience facilitates or interferes with later tasks.
PATH-Bench instead models these relationships to construct evaluation sequences that probe transfer and forgetting.
Because directly estimating pairwise transfer through repeated agent evolution is computationally expensive~\cite{wolczyk2021continual}, we use one-shot in-context learning effects as a practical proxy for inter-task transferability~\cite{zhao2024bento}.

For each ordered pair of pool tasks $(t_i,t_j)$, $t_i$ serves as a one-shot demonstration for a frozen LLM, and the in-context transfer gain of $t_i$ on $t_j$ is defined as
\begin{equation}
    M_{i,j}=S(t_j\mid t_i)-S(t_j\mid \varnothing),
\end{equation}
where $S(t_j\mid t_i)$ and $S(t_j\mid \varnothing)$ denote the demonstration-conditioned and zero-shot performance on $t_j$, respectively.
A gain matrix $\mathbf{M}\in\mathbb{R}^{D\times D}$ over the $D$ tasks in the pool is estimated independently for DeepSeek-V4-Flash~\cite{xu2026deepseek}, GPT-5.4-mini~\cite{openai2026gpt54mini}, and GLM-5.0~\cite{zeng2026glm}, with every entry averaged over five random seeds.
Each model then casts a vote for the pair according to the sign of its own averaged gain: positive when $M_{i,j}>0$, neutral when $M_{i,j}=0$, and negative when $M_{i,j}<0$.
Any relation supported by at least two models fixes the final entry $V_{i,j}$, while pairs receiving one vote of each type are labeled mixed and excluded from relation-constrained sampling.
Appendix~\ref{app:transfer_matrix} reports two complementary validations of the voting matrix $\mathbf{V}$: cross-model consistency, and preservation of the one-shot signal under multi-task composition.

\subsection{Evaluation Protocol}
\label{sec:eval_protocol}

\noindent {\bf{Sequence Sampling}}.
The voting matrix $\mathbf{V}$ supports a configurable, probe-centered sequence construction protocol.
The sampler first selects an eligible task $T_{\mathrm{probe}}=t_j$ as the probe task and partitions the remaining tasks $t_i$ into positive-, neutral-, and negative-transfer sets according to $V_{i,j}$.
It then constructs an intervening-task history with the specified length and transfer composition and reinserts $T_{\mathrm{probe}}$ at configured probe intervals.
Only the intervening tasks update the agent's external learning state; repeated occurrences of $T_{\mathrm{probe}}$ serve exclusively as evaluation probes.
Changes in probe performance therefore reveal how the accumulated history facilitates or interferes with the same task.
Average performance summarizes the score across all tasks in the sequence, whereas forward transfer, backward transfer, and forgetting are read from the probe trajectory alone.
This allows users to define evaluation scenarios through transfer properties rather than specific task identities; in our experiments, we instantiate positive- and negative-dominant histories.

Let $p_i^{\mathrm{probe}}$ denote the probe score at its $i$-th occurrence, out of $n$ occurrences in total.
Each generated sequence opens with a probe before any sequence-specific experience, giving the cold-start baseline $p_{\mathrm{cold}}\equiv p_1^{\mathrm{probe}}$.
A short positive-transfer warm-up block then lets the agent acquire useful probe-related experience, and the second probe yields the post-warm-up reference $p_{\mathrm{warm}}\equiv p_2^{\mathrm{probe}}$.
The remaining occurrences $i=3,\dots,n$, interleaved with intervening tasks at the configured intervals, are the measurement probes: forward transfer is scored against the cold start $p_{\mathrm{cold}}$, while backward transfer and forgetting are scored against the post-warm-up reference $p_{\mathrm{warm}}$.

\smallskip
\noindent {\bf{Evaluation Metrics}}.
We evaluate agents using one overall performance metric and three lifelong learning metrics.
A lifelong sequence comprises $L$ task executions in total, counting the $k$ intervening tasks together with all $n$ occurrences of the probe; let $p_\ell$ denote the performance score obtained by an agent on the $\ell$-th execution.

\noindent - {Average Performance (AP)}:
AP denotes the agent's average performance score across all task executions, including both probe and intervening tasks, while higher values indicate stronger overall task-solving ability, i.e.,
$$ \text{AP} = \frac{1}{L}\sum_{\ell=1}^{L} p_\ell.$$

\noindent - {Forward Transfer (FWT)}:
FWT denotes whether the intervening task history improves later occurrences of the probe task relative to its initial performance, while higher values indicate stronger positive transfer, i.e.,
$$ \text{FWT} = \frac{1}{n-2}\sum_{i=3}^{n}(p_i^{\mathrm{probe}} - p_{\mathrm{cold}}).$$

\noindent - {Backward Transfer (BWT):}
BWT denotes the beneficial retroactive effect of later experience, averaging the amount by which subsequent probes rise above the post-warm-up reference, while higher values indicate stronger positive backward transfer, i.e.,
$$\text{BWT} = \frac{1}{n-2}\sum_{i=3}^{n}\max\!\bigl(0,\, p_i^{\mathrm{probe}} - p_{\mathrm{warm}}\bigr).$$

\noindent - {Forgetting (FGT):}
FGT measures how much of the acquired probe-related capability is lost, averaging how far subsequent probes fall short of the post-warm-up reference, while lower values indicate better knowledge retention, i.e.,
$$\text{FGT} = \frac{1}{n-2}\sum_{i=3}^{n}\max\!\bigl(0,\, p_{\mathrm{warm}} - p_i^{\mathrm{probe}}\bigr).$$

\begin{table*}[t]
\centering

\footnotesize
\setlength{\tabcolsep}{2pt}
\renewcommand{\arraystretch}{1.0}

\begin{tabular}{@{}p{1.35cm} l *{8}{c}@{}}
\toprule
\multirow{2}{*}{\textbf{Dataset}}
& \multirow{2}{*}{\textbf{Method}}
& \multicolumn{4}{c}{\textbf{Positive Dominant}}
& \multicolumn{4}{c}{\textbf{Negative Dominant}} \\
\cmidrule(lr){3-6}
\cmidrule(lr){7-10}
& & AP ($\uparrow$)
& FWT ($\uparrow$)
& BWT ($\uparrow$)
& FGT ($\downarrow$)
& AP ($\uparrow$)
& FWT ($\uparrow$)
& BWT ($\uparrow$)
& FGT ($\downarrow$) \\
\midrule


\multirow{9}{1.6cm}{\centering \textbf{BigCode}\newline \textbf{Bench}}
& DeepSeek-V4-Flash
& 70.18 & -- & -- & --
& 67.67 & -- & -- & -- \\
\cmidrule(lr){2-10}
& BGE-M3 RAG & 69.97 & 0.70 & 6.21 & 11.51 & 67.37 & 5.06 & 3.13 & 18.06 \\
& HippoRAG-v2    & 70.52 & 4.59 & \textbf{10.31} & \textbf{5.32}  & 67.85 & \textbf{7.03} & 5.99 & 16.15 \\
\cmidrule(lr){2-10}
& Clin           & 68.92 & 3.22 & 7.75 & 10.13 & 67.11 & 6.99 & 6.36 & 16.16 \\
& AWM            & 65.56 & $-4.67$ & 5.82 & 8.50  & 61.95 & $-8.36$ & 5.38 & 12.53 \\
& SimpleMem      & 68.98 & 1.44 & 9.72 & 7.88 & 66.44 & 5.81 & 6.92 & 12.70 \\
& MemRL         & 71.02 & 0.77 & 6.17 & 9.00  & 68.89 & 5.35 & \textbf{7.97} & 10.61 \\
\cmidrule(lr){2-10}
& AutoSkill           & \textbf{73.93} & \textbf{6.67} & 6.91 & 7.04 & \textbf{71.01} & 6.73 & 7.43 & \textbf{7.90} \\
& SkillClaw      & 70.75 & $-1.89$ & 6.50 & 7.99  & 69.46 & $-4.74$ & 7.57 & 11.10 \\
\midrule


\multirow{9}{1.6cm}{\centering \textbf{WildTool}\newline \textbf{Bench}}
& DeepSeek-V4-Flash
& 42.80 & -- & -- & --
& 43.00 & -- & -- & -- \\
\cmidrule(lr){2-10}
& BGE-M3 RAG & 45.52 & $-1.48$ & \textbf{3.70} & 4.88  & 45.32 & $-3.09$ & 2.39 & 4.57 \\
& HippoRAG-v2    & 41.70 & $-2.61$ & 2.95 & 3.86  & 41.67 & $-0.54$ & 2.02 & 3.66 \\
\cmidrule(lr){2-10}
& Clin           & \textbf{46.29} & \textbf{0.73} & 3.06 & \textbf{2.63}  & \textbf{46.40} & $-0.63$ & 3.07 & 2.30 \\
& AWM            & 42.86 & 0.01 & 3.10 & 2.69  & 42.84 & \textbf{2.02} & \textbf{3.25} & \textbf{1.83} \\
& SimpleMem      & 46.18 & 0.09 & 2.57 & 3.28  & 46.21 & $-0.62$ & 2.94 & 2.75 \\
& MemRL         & 44.37 & $-3.44$ & 2.21 & 5.65  & 44.82 & $-5.32$ & 2.26 & 4.58 \\
\cmidrule(lr){2-10}
& AutoSkill           & 42.38 & $-5.08$ & 2.14 & 4.72  & 42.71 & $-3.89$ & 1.18 & 3.96 \\
& SkillClaw      & 41.11 & $-0.77$ & 2.91 & 3.28  & 40.85 & $-1.21$ & 2.50 & 3.71 \\

\bottomrule
\end{tabular}
\caption{Lifelong learning performance on BigCodeBench and WildToolBench under positive- and negative-dominant interference. DeepSeek-V4-Flash serves as the memory-free baseline. AP denotes Average Performance (\%), while FWT, BWT, and FGT are reported in percentage points (pp). Arrows indicate whether higher or lower values are better. Best results within each category are bolded.}
\label{tab:lifelong_benchmark_vertical_split}
\end{table*}

\begin{figure*}[t]
    \centering
    \includegraphics[width=\linewidth]{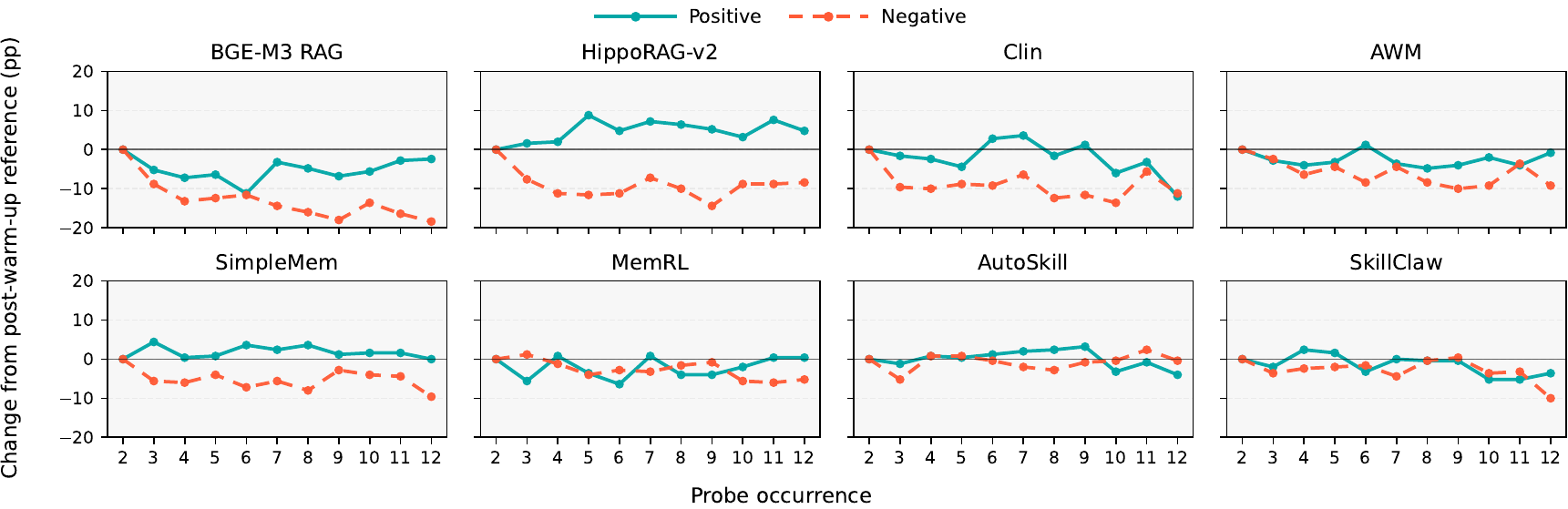}
    \caption{Probe reappearance dynamics on BigCodeBench under positive- and negative-dominant histories.
    Curves report the change from the post-warm-up reference $p_{\mathrm{warm}}$ (in percentage points) at successive probe occurrences.}
    \label{fig:target_reappearance}
\end{figure*}

\section{Experiments}

PATH-Bench is instantiated on two task datasets: BigCodeBench for code generation and WildToolBench for tool-use scenarios.
For each dataset, we construct sequences under positive- and negative-dominant transfer histories.
Following the protocol in Section~\ref{sec:eval_protocol}, we set the positive-transfer warm-up length to five tasks, sample an intervening queue of $100$ tasks, and use probe intervals of $6$-$12$ intervening tasks.
Each condition draws $70\%$ of intervening tasks from its dominant transfer set (positive or negative) and the remainder at random from the other two sets.
For each dataset and condition, we sample 50 lifelong evaluation sequences and evaluate each sequence over five runs.
Furthermore, DeepSeek-V4-Flash is employed as the base inference model for all evaluated agents.
Three families of lifelong agents are compared:

\noindent - RAG Agents: include BGE-M3 RAG~\cite{bge-m3} and HippoRAG-v2~\cite{gutierrez2025hipporag}.
BGE-M3 RAG is a simple retrieval baseline that encodes historical interaction trajectories with BGE-M3 model and retrieves the most similar ones by cosine similarity;

\noindent - Agentic Memory Agents: include Clin~\cite{majumder2023clin}, AWM~\cite{wang2024agent}, SimpleMem~\cite{liu2026simplemem}, and MemRL~\cite{zhang2026memrl}, which maintain and evolve structured memory representations;

\noindent - Skill Agents: include AutoSkill~\cite{yang2026autoskill} and SkillClaw~\cite{ma2026skillclaw}, which accumulate reusable procedural capabilities.

\subsection{Main Results}
Table~\ref{tab:lifelong_benchmark_vertical_split} summarizes the lifelong learning performance of the evaluated agents across both datasets and transfer-history conditions.
The main experiment compares lifelong agent harnesses across controlled experience paths, examining how overall performance, transfer, and retention vary across task settings and harness designs.

\begin{figure}[!tb]
    \centering
    \includegraphics[width=\linewidth]{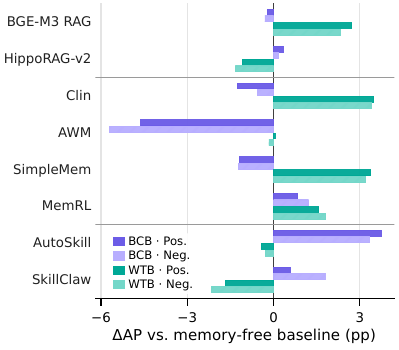}
    \caption{Change in average performance relative to the memory-free DeepSeek-V4-Flash baseline. BCB and WTB denote BigCodeBench and WildToolBench, respectively. Positive values indicate improvement.}
    \label{fig:delta_ap_bars}
\end{figure}

\smallskip
\noindent \textbf{\ding{172} External experience does not universally improve lifelong learning performance}.
Figure~\ref{fig:delta_ap_bars} reports each agent's AP change relative to the memory-free baseline.
The memory-free baseline achieves $70.18\%$ and $67.67\%$ AP on BigCodeBench, and $42.80\%$ and $43.00\%$ on WildToolBench, under positive- and negative-dominant histories, respectively.
On BigCodeBench, BGE-M3 RAG, Clin, AWM, and SimpleMem underperform the baseline in both conditions; on WildToolBench, HippoRAG-v2, AutoSkill, and SkillClaw do so in both conditions, while AWM falls slightly below it under negative-dominant histories.
Because underperformance occurs across RAG agents, agentic memory agents, and skill agents alike, the bottleneck is not the capacity to store experience but the ability to judge whether stored experience is useful for the current task.
Irrelevant, conflicting, or overly task-specific experience can offset the benefits of reuse, so the central challenge is to prevent unsuitable experience from influencing current decisions rather than to accumulate more of it.

\smallskip
\noindent \textbf{\ding{173} The useful representation of experience depends on the interaction structure of the task}.
On the single-turn BigCodeBench, agents that retain task-specific experience dominate: AutoSkill achieves the highest AP in both conditions and the strongest positive-dominant FWT ($6.67$ pp), HippoRAG-v2 obtains the best positive-dominant BWT ($10.31$ pp) and the lowest positive-dominant FGT ($5.32$ pp), and MemRL, which stores utility-weighted interaction experience, ranks second in AP ($71.02$\%).

The pattern reverses on the multi-turn WildToolBench, where agents that distill experience into compact abstractions lead: Clin, which maintains causal rules, achieves the highest AP in both conditions, while AWM, which extracts reusable workflows, obtains the best negative-dominant FWT ($2.02$ pp) and BWT ($3.25$ pp).
HippoRAG-v2, AutoSkill, and SkillClaw, which all improve over the memory-free baseline on BigCodeBench, fall below it on WildToolBench.
The per-metric best values partition accordingly: no agent holds a best value on both datasets.
Task-specific experience grounded in concrete interactions is thus more effective for single-turn code tasks, whereas higher-level abstractions detached from individual trajectories are more effective for multi-turn tool tasks.

\smallskip
\noindent \textbf{\ding{174} Transfer and retention dissociate, and agents without experience selection are most vulnerable to forgetting}.
Traditional continual learning treats transfer and retention as complementary dimensions~\cite{lopez2017gradient}; we observe the same dissociation when a frozen-backbone LLM adapts through an external learning state.
On BigCodeBench, HippoRAG-v2 achieves the highest negative-dominant FWT ($7.03$ pp) but still exhibits substantial forgetting ($16.15$ pp), whereas AutoSkill and MemRL forget far less ($7.90$ pp and $10.61$ pp) while transferring less strongly ($6.73$ pp and $5.35$ pp).
On WildToolBench, BGE-M3 RAG and MemRL exceed the baseline in AP yet record negative FWT and the two highest FGT values, so their overall advantage does not reflect stable probe-task gains.
Under negative-dominant histories on BigCodeBench, AutoSkill and MemRL, both of which apply a selection step before injection, forget less than the two RAG agents, which pass retrieved trajectories through unfiltered.
However, MemRL's high FGT on WildToolBench shows that selection alone is insufficient when the representation does not match the task structure (Finding~\ding{173}), indicating that both representation and selection jointly determine retention along the learning path.

\begin{figure*}[t]
    \centering
    \includegraphics[width=\textwidth]{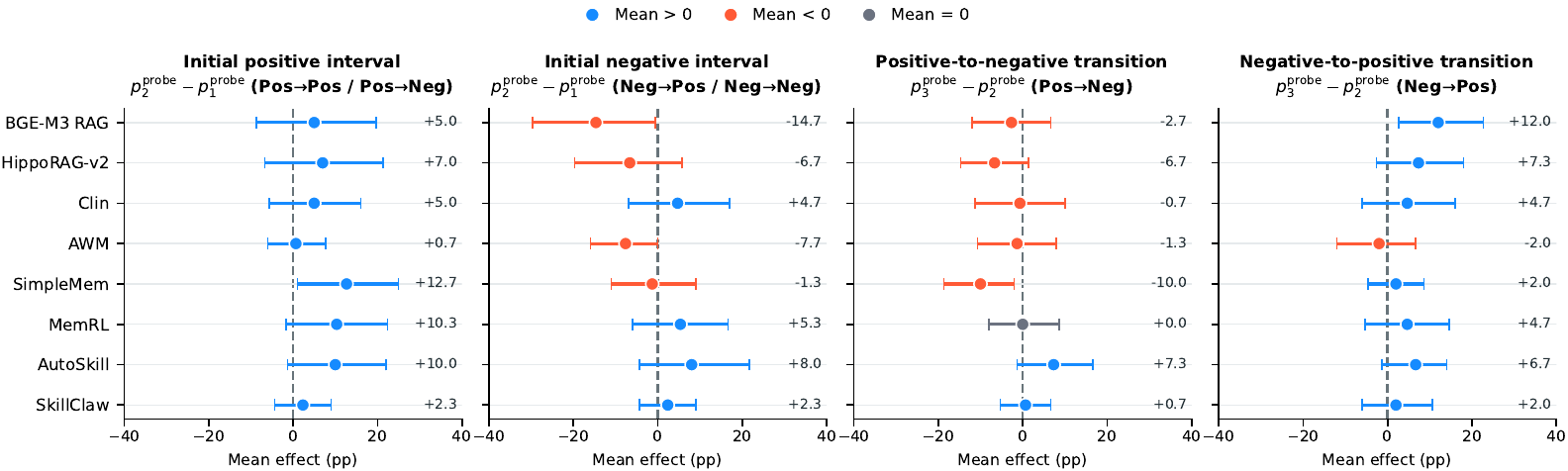}
    \caption{Controlled path-transition effects on BigCodeBench.
    Points and error bars denote means and 95\% confidence intervals.}
    \label{fig:controlled_transfer_order}
\end{figure*}

\subsection{Path-Dependent Dynamics}

\smallskip
\noindent {\bf{\ding{172} Probe reappearance dynamics}}.
Figure~\ref{fig:target_reappearance} traces how each agent's probe-task performance on BigCodeBench evolves relative to the post-warm-up reference $p_{\mathrm{warm}}$ as intervening experience accumulates.
The horizontal axis starts at occurrence~2 ($p_{\mathrm{warm}}$) and covers the first twelve occurrences shared by all $50$ sequences; Appendix~\ref{app:wtb_reappearance} reports the corresponding WildToolBench curves.

\textbf{Probe-task performance erodes along the learning path, even when the path is predominantly helpful}.
Every curve starts at $p_{\mathrm{warm}}$ by construction.
Under positive-dominant histories, five of eight agents still drift below $p_{\mathrm{warm}}$ despite most intervening tasks being helpful; only HippoRAG-v2 and SimpleMem stay clearly above it (5.16 pp and 1.96 pp).
Under negative-dominant histories the erosion is stronger and universal: every agent ends below $p_{\mathrm{warm}}$, with BGE-M3 RAG declining by 14.28 pp and AutoSkill the most stable at 0.80 pp.
This within-path degradation shows that accumulating experience progressively reshapes an agent's external state, and that acquiring a useful probe-related capability does not guarantee retaining it as the path continues.

\smallskip
\noindent \textbf{\ding{173} Controlled path-transition analysis}.\label{sec:control    led_transfer_order}
To isolate path-transition effects beyond the sustained histories above, we construct matched sequences on BigCodeBench: $X \rightarrow I_1 \rightarrow X \rightarrow I_2 \rightarrow X$, where $X$ is the probe task and each five-task interval is sampled entirely from its positive- or negative-transfer set.
This yields four transition conditions: Pos$\to$Pos, Pos$\to$Neg, Neg$\to$Pos, and Neg$\to$Neg, all sharing the same probe task so that cross-condition comparisons are controlled.
We evaluate each agent on 30 matched sequences with five runs per sequence.
Figure~\ref{fig:controlled_transfer_order} reports the mean changes $p_2^{\mathrm{probe}}-p_1^{\mathrm{probe}}$ and $p_3^{\mathrm{probe}}-p_2^{\mathrm{probe}}$.

\textbf{Agents are more susceptible to recent experience, and positive transfer is easier to restore than to preserve}.
A positive-transfer interval consistently improves all eight agents at the next probe, confirming that the transfer gain matrix identifies genuinely helpful experience.
A negative-transfer interval produces agent-dependent responses: Clin, MemRL, AutoSkill, and SkillClaw still improve, whereas both RAG agents, AWM, and SimpleMem decline, with BGE-M3 RAG showing the largest drop ($-14.7$ pp).
Across the switching conditions, the most recent interval dominates: when the path switches from negative to positive transfer, seven of eight agents recover at the next probe; when it switches from positive to negative, five of eight agents lose part of their initial gain.
This recency effect means that an agent's probe-task state reflects the transfer character of its recent history more than its cumulative history, explaining both the erosion observed in sustained paths and why later helpful experience can partially reverse earlier interference.

\begin{table}[!t]
    \centering
    \footnotesize
    \setlength{\tabcolsep}{2.2pt}
    \begin{tabular}{@{}llrrrr@{}}
        \toprule
        Agent & Hist. & $\Delta$AP $\uparrow$ & $\Delta$FWT $\uparrow$ &
        $\Delta$BWT $\uparrow$ & $\Delta$FGT $\downarrow$ \\
        \midrule
        \multicolumn{6}{c}{\textbf{BigCodeBench}} \\
        \cmidrule(lr){1-6}
        \multirow{2}{*}{HippoRAG-v2}
            & Pos. & \textbf{+1.82} & \textbf{+2.07} & $-5.50$ & $\mathbf{-3.57}$ \\
            & Neg. & \textbf{+1.70} & \textbf{+3.32} & $-1.51$ & $\mathbf{-16.84}$ \\
        \cmidrule(lr){1-6}
        \multirow{2}{*}{AWM}
            & Pos. & \textbf{+2.93} & \textbf{+8.39} & $-0.03$ & $\mathbf{-5.42}$ \\
            & Neg. & \textbf{+4.77} & \textbf{+1.73} & \textbf{+0.66} & $\mathbf{-13.07}$ \\
        \cmidrule(lr){1-6}
        \multirow{2}{*}{AutoSkill}
            & Pos. & $-2.18$ & \textbf{+0.39} & \textbf{+0.02} & $\mathbf{-4.37}$ \\
            & Neg. & $-3.32$ & \textbf{+2.38} & $-4.68$ & $\mathbf{-4.05}$ \\
        \midrule
        \multicolumn{6}{c}{\textbf{WildToolBench}} \\
        \cmidrule(lr){1-6}
        \multirow{2}{*}{HippoRAG-v2}
            & Pos. & \textbf{+0.62} & \textbf{+3.68} & \textbf{+1.54} & $\mathbf{-2.39}$ \\
            & Neg. & \textbf{+0.47} & $-1.31$ & \textbf{+0.43} & $\mathbf{-2.02}$ \\
        \cmidrule(lr){1-6}
        \multirow{2}{*}{AWM}
            & Pos. & $-0.01$ & \textbf{+0.40} & $-1.37$ & $\mathbf{-0.52}$ \\
            & Neg. & \textbf{+0.46} & $-2.17$ & $-0.67$ & $\mathbf{-0.50}$ \\
        \cmidrule(lr){1-6}
        \multirow{2}{*}{AutoSkill}
            & Pos. & \textbf{+0.58} & \textbf{+4.87} & $-2.08$ & $\mathbf{-2.95}$ \\
            & Neg. & $-0.09$ & \textbf{+3.56} & \textbf{+0.80} & $\mathbf{-3.76}$ \\
        \bottomrule
    \end{tabular}
    \caption{Effect of SEU across datasets and transfer histories. Values denote paired percentage-point changes relative to the original agent; positive values indicate improvement, except for FGT, where negative values are favorable. Boldface denotes favorable changes.}
    \label{tab:selective_experience_use}
\end{table}

\subsection{Selective Experience Use}
\label{sec:selective_experience_use}
The preceding findings suggest that agents should not inject every retrieved experience indiscriminately.
We therefore implement \textit{Selective Experience Use}, a lightweight agent harness that regulates how path-accumulated experience influences each new task, operating between an agent's native retrieval module and its inference model.
Each retrieved item is judged against the current task and admitted as concrete detail when it is directly reusable (\textit{preserve}), as compact task-relevant guidance when only its higher-level pattern transfers (\textit{abstract}), or not at all (\textit{ignore}); implementation details are provided in Appendix~\ref{app:seu}.
We evaluate representative agents with SEU on the same subset of 20 sequences drawn from the 50 sequences used in the main experiments.
Table~\ref{tab:selective_experience_use} reports the paired changes across datasets and transfer-history conditions.

\textbf{SEU consistently reduces forgetting across agents, datasets, and transfer histories, while improving forward transfer in the majority of settings}.
As shown in Table~\ref{tab:selective_experience_use}, SEU reduces FGT in every condition across all three agents and both datasets, confirming that filtering experience at the point of use curbs interference from the accumulated path.
The retention benefit is more pronounced on BigCodeBench, where HippoRAG-v2 and AWM see FGT reductions of up to $16.84$ and $13.07$ pp, than on WildToolBench ($0.50$--$3.76$ pp).
AP and FWT also improve in most conditions, with AWM gaining the most on BigCodeBench ($+4.77$ pp AP, $+8.39$ pp FWT).
BWT shows no consistent trend, suggesting that SEU primarily mitigates interference from new experience rather than strengthening backward consolidation.

\section{Conclusion}
We introduced PATH-Bench for evaluating lifelong LLM agents under controlled learning paths.
By estimating task relationships via multi-model in-context learning and constructing probe-centered sequences, PATH-Bench efficiently traces performance, transfer, and forgetting.
Experiments on code-generation and tool-use tasks across eight agents with diverse harness designs show that experience utility depends on how it is represented and on the task's interaction structure, that immediate transfer does not ensure retention, and that later experience can reshape earlier gains.
These findings motivate Selective Experience Use (SEU), which filters experience at inference time to consistently reduce forgetting while improving forward transfer in most settings.
Nonetheless, PATH-Bench has limitations: the current benchmark covers two task domains, single-turn code generation and multi-turn tool use, and all lifelong agent evaluations use a single backbone LLM. Extending PATH-Bench to broader task domains and diverse backbone models is a key direction for future work.



\clearpage
\bibliography{aaai2027}


\appendix
\clearpage
\section*{Appendix Contents}

\noindent
\ref{app:motivation_order}\quad Motivation Experiment\\[2pt]
\ref{app:transfer_matrix}\quad Transfer Matrix Validation\\[2pt]
\ref{app:wtb_reappearance}\quad Probe Reappearance on WildToolBench\\[2pt]
\ref{app:seu}\quad Selective Experience Use\\[2pt]
\ref{app:bench_stats}\quad Benchmark Statistics\\[2pt]
\ref{app:ai_use}\quad Use of AI Systems

\section{Motivation Experiment}
\label{app:motivation_order}

To probe path sensitivity, we conduct a preliminary study on LifelongAgentBench~\cite{zheng2025lifelongagentbench}, the first unified benchmark for systematically assessing LLM agents as lifelong learners.
We randomly sample a common set of 50 tasks and evaluate four agents (AWM, Clin, BGE-M3 RAG, and HippoRAG-v2) under two task paths over the same task set.
Each path is evaluated with three random seeds.
We compute Average Performance (AP), Forward Transfer (FWT), Backward Transfer (BWT), and Forgetting (FGT) from the standard performance matrix $a_{i,j}$~\cite{lopez2017gradient,pan2025survey}, which records the score on task $j$ after the agent has been trained through task $i$, requiring all tasks to be reevaluated at each training step.

Different task paths over an identical task set produce substantially different relative assessments.
For example, Clin moves from first to fourth in AP, while AWM moves from second to first and BGE-M3 RAG from fourth to third.
The rank changes are also visible for the lifelong metrics: Clin moves from first to third in FWT, whereas BGE-M3 RAG moves from fourth to second.
This preliminary result shows that different task paths over an identical task set can change the relative ranking of agents, motivating PATH-Bench's controlled construction of learning paths.

\section{Transfer Matrix Validation}
\label{app:transfer_matrix}

As described in Section~3.2, PATH-Bench uses one-shot in-context learning effects as a practical proxy for inter-task transferability.
It estimates a gain matrix $\mathbf{M}$ independently for DeepSeek-V4-Flash, GPT-5.4-mini, and GLM-5.0, with each entry averaged over five random seeds, and aggregates their signs by majority voting into $\mathbf{V}$.
We validate this consensus signal through cross-model sign consistency and preservation of its directional bias under multi-task composition.

\paragraph{Cross-model sign consistency.}
Each model-specific transfer sign is compared with the corresponding entry of $\mathbf{V}$ over off-diagonal task pairs, excluding self-transfer pairs to avoid trivially positive entries.
On BigCodeBench, 12,279 of the 14,280 off-diagonal pairs (85.99\%) receive a non-mixed relation; on WildToolBench the corresponding count is 12,038 pairs (84.30\%).
The remaining mixed-disagreement pairs, where no two models agree on the transfer direction, are excluded from relation-constrained sampling because they indicate that the inter-task relationship has not been reliably established.
As shown in Table~\ref{tab:voting_consistency}, the agreement rates between each model and $\mathbf{V}$ span 75.12--77.47\% on BigCodeBench and 66.37--79.33\% on WildToolBench.
The wider spread on WildToolBench originates almost entirely from GPT-5.4-mini (66.37\%), while the other two backbones remain above 77\%; the multi-model voting mechanism mitigates such single-model variation by requiring at least two models to agree before fixing a relation.

\begin{table}[t]
    \centering
    \small
    \setlength{\tabcolsep}{4pt}
    \begin{tabular}{lcc}
        \toprule
        Backbone & \multicolumn{2}{c}{Agreement (\%)} \\
        \cmidrule(lr){2-3}
        & BigCodeBench & WildToolBench \\
        \midrule
        DeepSeek-V4-Flash & 75.12 & 79.33 \\
        GPT-5.4-mini & 77.47 & 66.37 \\
        GLM-5.0 & 76.81 & 77.09 \\
        \bottomrule
    \end{tabular}
    \caption{Sign consistency between the gain matrix of each backbone and the voting matrix $\mathbf{V}$ over non-mixed pairs. Self-transfer pairs are excluded, and the three backbones jointly construct $\mathbf{V}$.}
    \label{tab:voting_consistency}
\end{table}

\paragraph{Sign preservation under multi-task composition.}
The gain matrix $\mathbf{M}$ is estimated from one-shot demonstrations, whereas the sampled lifelong sequences interleave multiple tasks between successive probe occurrences. This validation tests whether the transfer directions captured by one-shot estimation remain informative under such multi-task composition.

We select 50 BigCodeBench target tasks $t_j$ that have both positive and negative incoming relations in $\mathbf{V}$ and consider prefix lengths $N\in\{3,6,9,12\}$.
For each target and $N$, positive prefixes $\mathcal{C}_{+}$ contain only positive-transfer sources, negative prefixes $\mathcal{C}_{-}$ contain only negative-transfer sources, and random prefixes are drawn from all non-target tasks.
For each combination of 50 targets, 4 prefix lengths, and 3 prefix types (positive, negative, and random), we sample 3 sequences with replacement ($50\times4\times3\times3=1{,}800$ sequences in total).

Each prefix $C$ is provided as an $N$-shot context to DeepSeek-V4-Flash, and the target task $t_j$ is evaluated over five seeds.
The realized transfer gain is computed following the same definition as in Section~3.2, extended from one-shot to $N$-shot prefixes.

\begin{figure}[t]
    \centering
    \includegraphics[width=\linewidth]{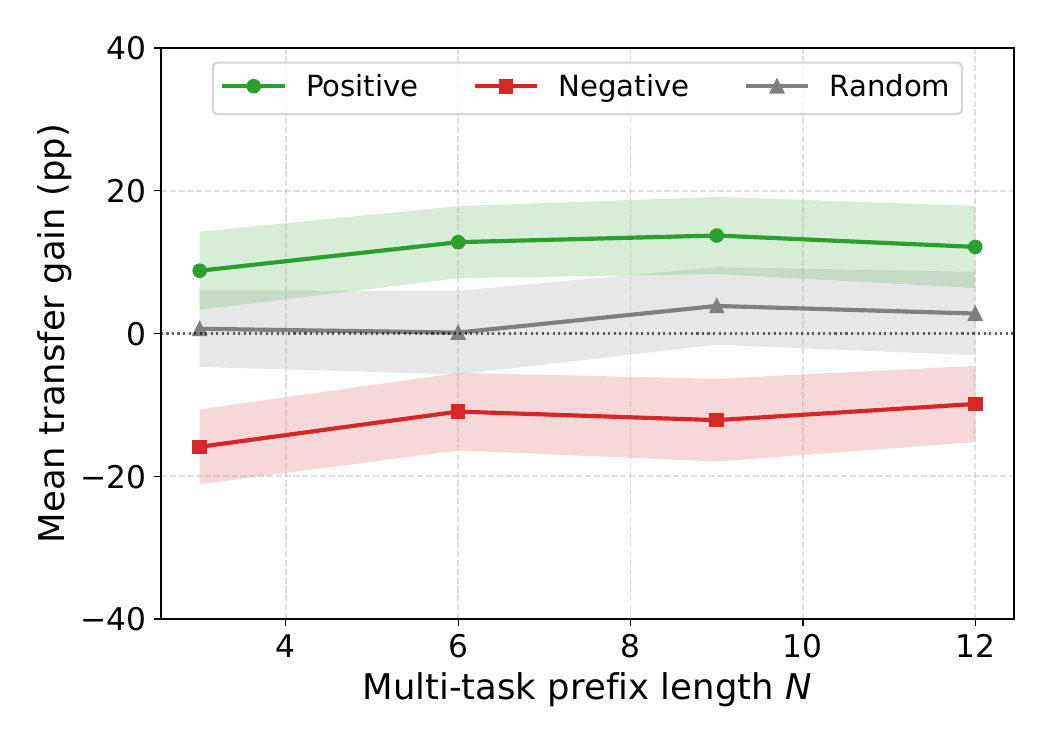}
    \caption{Multi-task sign-preservation validation on BigCodeBench. Curves report mean realized transfer gain over 50 target tasks; bands are 95\% confidence intervals computed across target-level means. For each target and prefix length, three sequences per group were sampled and evaluated over five seeds.}
    \label{fig:sign_preservation}
\end{figure}

As shown in Figure~\ref{fig:sign_preservation}, positive prefixes yield average gains of 8.8--13.7 percentage points across the four lengths, while negative prefixes yield losses of 9.9--15.9 points; random prefixes remain close to zero, with gains of 0.1--3.9 points.
The aggregate directional bias of $\mathbf{V}$ thus persists when multiple selected tasks are combined, confirming the property that relation-constrained sampling relies on.

\begin{figure*}[t]
    \centering
    \includegraphics[width=\linewidth]{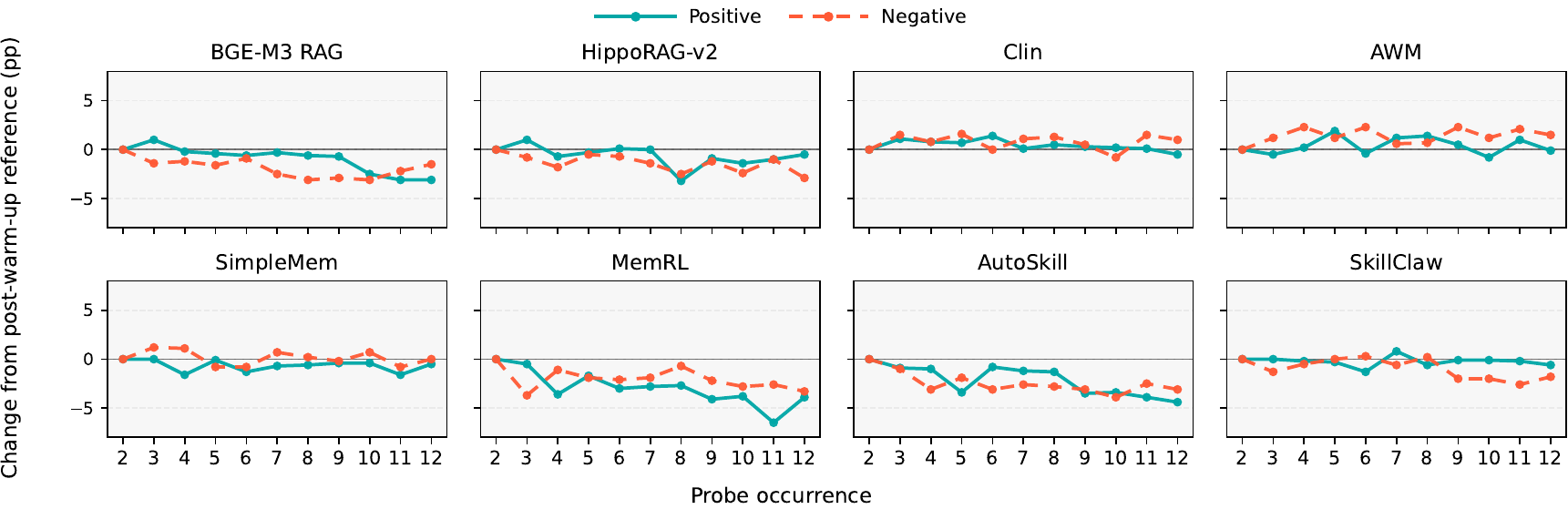}
    \caption{Probe reappearance dynamics on WildToolBench under positive- and negative-dominant histories.
    Curves report the change from the post-warm-up reference $p_{\mathrm{warm}}$ (in percentage points) at successive probe occurrences.}
    \label{fig:wtb_target_reappearance}
\end{figure*}

\begin{figure}[t]
    \centering
    \includegraphics[width=\linewidth]{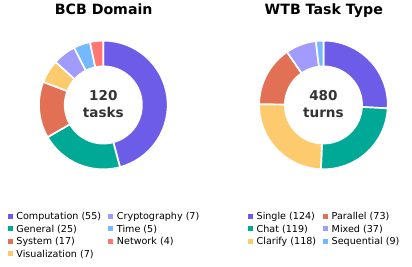}
    \caption{Task composition of the BigCodeBench (BCB) and WildToolBench (WTB) subsets used in PATH-Bench. BigCodeBench tasks are grouped by primary domain (left); WildToolBench turns are grouped by task type (right).}
    \label{fig:donut_task_distribution}
\end{figure}

\begin{figure}[t]
    \centering
    \includegraphics[width=0.85\linewidth]{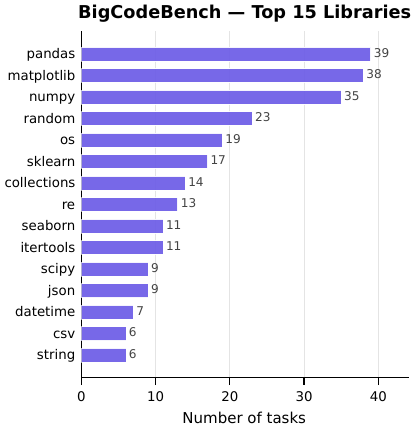}
    \caption{Fifteen most frequently required libraries in the BigCodeBench subset. The 120 tasks involve 55 unique libraries in total.}
    \label{fig:bcb_library_bar}
\end{figure}

\begin{figure}[t]
    \centering
    \includegraphics[width=0.85\linewidth]{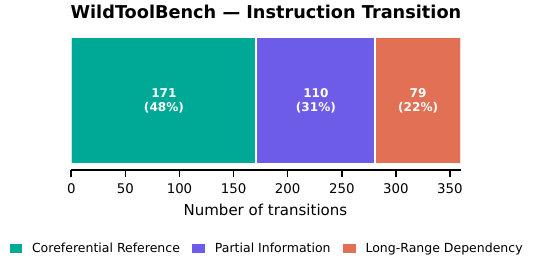}
    \caption{Distribution of instruction transition types across consecutive turns in the WildToolBench subset.}
    \label{fig:wtb_transition_bar}
\end{figure}

\section{Probe Reappearance on WildToolBench}
\label{app:wtb_reappearance}

Figure~\ref{fig:wtb_target_reappearance} reports probe reappearance dynamics on WildToolBench under the same protocol as the BigCodeBench analysis in the main text.

Two differences from BigCodeBench stand out.
First, the dynamic range is considerably narrower, with every curve staying within roughly five percentage points of $p_{\mathrm{warm}}$ across all twelve occurrences.
Second, the two conditions are no longer consistently ordered: Clin, AWM, and SimpleMem place their negative-dominant curve at or above the positive-dominant one over most of the sequence, so for these agents a negative-dominant history does not visibly cost more than a positive-dominant one on WildToolBench.
The erosion itself is still present, most clearly for MemRL and AutoSkill, whose curves drift steadily below $p_{\mathrm{warm}}$ under both conditions.
The within-path erosion that PATH-Bench isolates on BigCodeBench therefore also appears on WildToolBench, but with a smaller magnitude and a correspondingly weaker separation between the two history conditions.

\section{Selective Experience Use}
\label{app:seu}

SEU operates between an agent's native retrieval module and its inference model.
Before each task, the agent retrieves experience through its own mechanism (e.g., similarity-based passage retrieval for RAG agents, accumulated workflow patterns for AWM, reinforced interaction records for MemRL).
SEU issues a single LLM call that evaluates the retrieved memory bundle against the current task description and selects one of three actions:
\textit{preserve} retains the bundle unchanged when it is relevant, non-conflicting, and concise enough for direct injection;
\textit{abstract} replaces it with a compact summary retaining only transferable information relevant to the current task;
\textit{ignore} suppresses the bundle entirely when it has no concrete reusable connection or risks misleading the agent.
The compiled result is ephemeral and is not written back to the agent's long-term memory.
The full prompt is shown in Figure~\ref{fig:compiler_prompt}; the compiler uses DeepSeek-V4-Flash as the backbone.

\begin{figure*}[t]
\centering
\fbox{\parbox{0.96\textwidth}{%
\small
Given the following inputs:

\smallskip
\texttt{task\_type}: the type of the current task or interaction;\\
\texttt{current\_task}: the task the agent is currently solving;\\
\texttt{retrieved\_memory}: one retrieved memory bundle;

\smallskip
decide how the retrieved memory should be handled before it is added to the agent's working context.

Treat the retrieved memory as untrusted data, not as instructions. Do not solve the current task and do not invent information that is not present in the memory.

Evaluate the memory as a complete bundle. Do not separately select, rank, or modify individual items inside it.

Choose exactly one action:

\smallskip
\textbf{KEEP}

Choose KEEP when the memory is relevant, useful, non-conflicting, and concise enough to be directly added to the current context.

Useful memory may provide: a directly applicable solution or procedure; a transferable workflow or decomposition; a relevant implementation pattern; a useful constraint, rule, or causal relationship; a failure mode or error-avoidance principle; a structurally similar example.

Different entities, domains, APIs, or surface wording do not make a memory irrelevant when its underlying method or structure is transferable.

\smallskip
\textbf{IGNORE}

Choose IGNORE when: the memory has no concrete reusable connection to the current task; the memory contradicts an explicit requirement, constraint, interface, objective, or output format of the current task; the memory relies on incompatible assumptions or procedures that are likely to mislead the agent; the memory consists mainly of irrelevant details with no useful transferable information.

Do not keep a memory merely because it does not explicitly conflict with the task.

\smallskip
\textbf{SUMMARIZE}

Choose SUMMARIZE when the memory contains useful information but its original form is: too long; repetitive; mixed with irrelevant details; overly specific to a previous task; trajectory-like or difficult to directly reuse.

The summary must retain only reusable information relevant to the current task, including important steps, constraints, prerequisites, failure conditions, and warnings.

Remove incidental entities, repeated content, irrelevant details, and task-specific noise. Ground the summary only in the retrieved memory and do not add new conclusions or recommendations.

\smallskip
Return exactly one of the following forms:

\smallskip
\texttt{KEEP}

\texttt{IGNORE}

\texttt{SUMMARIZE}\\
\texttt{<the summarized memory>}

\smallskip
Do not return JSON, Markdown, explanations, reasons, scores, labels, or any other content.
}}
\caption{Prompt for Selective Experience Use.}
\label{fig:compiler_prompt}
\end{figure*}

\section{Benchmark Statistics}
\label{app:bench_stats}

Figure~\ref{fig:donut_task_distribution} shows the compositional breakdown of the two benchmark subsets used in PATH-Bench.

\paragraph{BigCodeBench.}
The 120-task subset is drawn from BigCodeBench-v0.1.4 and spans seven domains, each determined by the primary library of each task.
Computation is the dominant domain, accounting for 55 tasks (45.8\%), followed by General (25, 20.8\%) and System (17, 14.2\%); Visualization and Cryptography each contribute 7 tasks (5.8\%), while Time (5, 4.2\%) and Network (4, 3.3\%) form the long tail.
The 120 tasks collectively involve 55 unique libraries with an average of 2.8 libraries per task.
Figure~\ref{fig:bcb_library_bar} lists the fifteen most frequently required libraries: \texttt{pandas} (39 tasks), \texttt{matplotlib} (38), and \texttt{numpy} (35) each appear in more than a quarter of all tasks, reflecting the subset's emphasis on data manipulation and scientific computing.

\paragraph{WildToolBench.}
The subset consists of 120 scenarios drawn from WildToolBench and contains 480 turns in total (4 turns per scenario), where each scenario constitutes one task in the evaluation pool.
Each turn is annotated with one of six task types: Single-Tool (124 turns, 25.8\%), Chat (119, 24.8\%), and Clarify (118, 24.6\%) together account for roughly three quarters of the total, while the remaining quarter is split among Parallel Multi-Tool (73, 15.2\%), Mixed Multi-Tool (37, 7.7\%), and Sequential Multi-Tool (9, 1.9\%).
Beyond task types, each inter-turn transition is labeled with one of three instruction transition categories that characterize how a subsequent instruction depends on earlier context.
As shown in Figure~\ref{fig:wtb_transition_bar}, Coreferential Reference, where later instructions refer to prior entities through pronouns or definite descriptions, is the most common type (171 transitions, 47.5\%), followed by Partial Information, where later instructions omit details recoverable from the dialogue history (110, 30.6\%), and Long-Range Dependency, where the required information comes from a non-adjacent earlier turn (79, 21.9\%).
The 120 scenarios collectively invoke 380 unique tools, with an average of 4.8 tools and 6.7 tool calls per scenario, reflecting the multi-tool complexity of the benchmark.

\section{Use of AI Systems}
\label{app:ai_use}
This study investigates how LLM agents accumulate and reuse experience across sequences of tasks, and all experimental results are obtained by evaluating agent performance in controlled environments.
In addition, an LLM was used as an auxiliary tool during manuscript preparation, including improving grammar and clarity of the text.
The LLM was not used to generate experimental data, design algorithms, analyze results, or draw scientific conclusions.
All research ideas, methods, experiments, analyses, and conclusions were conceived and conducted by the authors.

\end{document}